\documentclass{article}
\usepackage[utf8]{inputenc}
\usepackage{thumbpdf,lmodern}
\usepackage[ruled,vlined]{algorithm2e}
\usepackage{amsmath}
\usepackage{framed}
\usepackage{graphicx,subcaption}
\usepackage{rotating}
\usepackage{authblk}
\usepackage{fancyvrb}
\usepackage{url}
\usepackage{hyperref}
\usepackage{setspace}
\usepackage{caption}
\usepackage{bbold}
\usepackage{multirow}
\usepackage{pdfpages}
\usepackage{xr}
\usepackage[%
  autocite    = superscript,
  backend     = bibtex,
  sorting   = none,
  uniquename = false,
  style       = nature,
  ]{biblatex}
  
 \renewcommand\cite{\autocite}  

\DeclareLanguageMapping{english}{english-apa}
\DeclareFieldFormat{labelnumberwidth}{\mkbibbrackets{#1}}
\defbibenvironment{bibliography}
  {\list
     {\printtext[labelnumberwidth]{%
      \printfield{labelprefix}%
      \printfield{labelnumber}}}
     {\setlength{\labelwidth}{\labelnumberwidth}%
      \setlength{\leftmargin}{\labelwidth}%
      \setlength{\labelsep}{\biblabelsep}%
      \addtolength{\leftmargin}{\labelsep}%
      \setlength{\itemsep}{\bibitemsep}%
      \setlength{\parsep}{\bibparsep}}%
      }
  {\endlist}
  {\item}

\usepackage{geometry}
\let\code=\texttt
\let\proglang=\textsf
\newcommand{\pkg}[1]{{\fontseries{b}\selectfont #1}}

\newcommand{\fct}[1]{\code{#1()}}

\DeclareMathOperator{\Avg}{Avg}
\DeclareMathOperator{\ACC}{ACC}
\DeclareMathOperator{\argmin}{argmin}
\newcommand{\E}{\mathsf{E}}

\DefineVerbatimEnvironment{Code}{Verbatim}{}
\DefineVerbatimEnvironment{CodeInput}{Verbatim}{fontshape=sl}
\DefineVerbatimEnvironment{CodeOutput}{Verbatim}{}

\def\@fnsymbol#1{\ensuremath{\ifcase#1\or \dagger\or \ddagger\or
   \mathsection\or \mathparagraph\or \|\or **\or \dagger\dagger
   \or \ddagger\ddagger \else\@ctrerr\fi}}

\title{\pkg{reval}: a \proglang{Python} package to determine best clustering solutions with stability-based relative clustering validation}

\author[1]{Isotta Landi\thanks{Correspondence: landi.isotta@gmail.com, @IsottaLandi}}
\author[1,2]{Veronica Mandelli}
\author[1,3]{Michael V. Lombardo}
\affil[1]{
Laboratory for Autism and Neurodevelopmental Disorders, Center for Neuroscience and Cognitive Systems @UniTn, Istituto Italiano di Tecnologia, Rovereto, Italy.
}
\affil[2]{
Center for Mind/Brain Sciences, University of Trento, Rovereto, Italy.
}
\affil[3]{
Autism Research Centre, Department of Psychiatry, University of Cambridge, Cambridge, United Kingdom.
}
\date{}
\providecommand{\keywords}[1]{\textbf{\textit{Keywords---}} #1}
\begin{document}
\maketitle

\begin{abstract}
Determining the best partition for a dataset can be a challenging task because of 1) the lack of a priori information within an unsupervised learning framework; and 2) the absence of a unique clustering validation approach to evaluate clustering solutions. Here we present \pkg{reval}: a \proglang{Python} package that leverages stability-based relative clustering validation methods to determine best clustering solutions as the ones that best generalize to unseen data. 
    
Statistical software, both in \proglang{R} and \proglang{Python}, usually rely on internal validation metrics, such as \emph{silhouette}, to select the number of clusters that best fits the data. Meanwhile, open-source software solutions that easily implement relative clustering techniques are lacking. Internal validation methods exploit characteristics of the data itself to produce a result, whereas relative approaches attempt to leverage the unknown underlying distribution of data points looking for generalizable and replicable results.

The implementation of relative validation methods can further the theory of clustering by enriching the already available methods that can be used to investigate clustering results in different situations and for different data distributions. This work aims at contributing to this effort by developing a stability-based method that selects the best clustering solution as the one that replicates, via supervised learning, on unseen subsets of data. The package works with multiple clustering and classification algorithms, hence allowing both the automatization of the labeling process and the assessment of the stability of different clustering mechanisms.
\end{abstract}

\section*{Introduction} 
\label{sec:intro}

Clustering algorithms identify intrinsic subgroups in a dataset by arranging together elements that show smaller pairwise dissimilarities relative to other subgroups \cite{stat}. They are one of a number of machine learning methods that do unsupervised learning with the aim of identifying patterns in the data in the absence of supervised/external knowledge. While their usage is relatively widespread, the lack of a priori information complicates the evaluation of clustering solutions. Attempts to address this challenge usually rely on internal validation measures, that focus on quantities and features inherent to a grouping solution \cite{vazirgiannis2009}. Many tools are available to compute internal validation measures that help in determining the best number of clusters. For example, the elbow method \cite{thorndike1953} selects the number of clusters for which the within-cluster variability decrement is minimal. Another popular method using internal criteria is the silhouette-based approach \cite{rousseeuw1987}. This method maximizes cluster cohesion and separation - that is, how similar an object is to other elements of the same cluster compared to elements of other clusters. 

In contrast to internal validation, other approaches, such as relative validation methods, have the potential to transform cluster analysis into a model selection problem and help evaluate the best clustering solution (i.e., best number of clusters). The way these methods are conceived also offer up the possibility to determine the extent to which a clustering solution generalizes to unseen data and hence to enable the replication of the data partition chosen. While a variety of software packages contain internal cluster validation methods and measures, open-source software to easily implement the full potential of relative validation techniques are lacking. Here we present the \pkg{reval} \proglang{Python} package (pronounced ``reh-val'', like the word ``revel'') to help fill this important missing gap. \pkg{reval} implements an approach for stability-based validation of clustering solutions described by Lange and colleagues \cite{lange2004} which allows for the identification and evaluation of partitions that best generalize to unseen data and the automation of the labeling process.

Stability-based methods return the number of clusters that minimizes the expected distance between clustering solutions obtained for different datasets. Several options are available \cite{luxburg} to 1) generate the datasets (e.g., random subsampling of the original dataset \cite{ben2001}, or adding random noise \cite{moller2006}); 2) compare clustering solutions (e.g., overlapping subsamples \cite{ben2001}); and 3) compute clustering distances (e.g., the consensus index by Vinh and Epps \cite{vinh2009}). The method proposed by Lange et al. \cite{lange2004} has the advantage of transforming the unsupervised setting into a classification problem and guides selection through the minimization of prediction error. First, a dataset is split into training and validation sets and then independently partitioned into clusters. Second, training set labels are used within supervised classification methods to learn how to best predict the labels. Applying the classification model to the validation set, the model’s predicted labels are then compared to the actual clustering labels derived from the validation set. This procedure is repeated using cross-validation and the optimal number of clusters is identified corresponding to the maximum number of clusters that minimizes prediction error. Prediction performance is defined by the authors as the 0-1 loss in supervised classification \cite{lange2004, luxburg}, namely, the normalized Hamming distance. Nevertheless, other choices are possible, for example, Tibshirani and colleagues used prediction strength - that is, the proportion of observation pairs in the validation set that are assigned to the same cluster by both the clustering algorithm and the partition based on the training set centers \cite{tibshirani2005}. 

Libraries and methods for the automated selection of the best number of clusters are available in both \proglang{Python} and \proglang{R}. The \pkg{yellowbrick} \proglang{Python} visual analysis and diagnostic tool suite \cite{yellowbrick2019} includes the implementation of the elbow method to determine the best number of clusters. In \proglang{R}, \pkg{NbClust} \cite{nbclust} is a popular library that compiles 30 different internal metrics and allows for users to compute all or a subset of these metrics for use in a majority vote rule to select the optimal number of clusters. For relative validation approaches there are the \pkg{clValid} \cite{clvalid} and \pkg{cstab} \cite{cstab} libraries that apply stability-based relative validation models. \pkg{clValid} was designed to work with highly correlated high-throughput genomic data and computes stability measures comparing clustering solutions based on full data and data with a single column removed. \pkg{cstab} implements the selection of the best number of clusters via model-based and model-free clustering instability \cite{haslbeck2016} using a bootstrap approach.

The \pkg{reval} package contributes to this landscape by implementing a stability-based approach that can be easily applied to different datasets using multiple clustering and classification algorithms. Built on top of the stability-based algorithm \cite{lange2004}, \pkg{reval} applies a classifier trained on the best clustering solution to a test set, returning classification metrics that help interpret the generalization performance, guide the clustering process, and enable labeling replication. Such a tool can be used in concert with internal measures to assess the underlying structure of a dataset to help avoid the risk of overfitting. With respect to clustering errors internal and relative indices can exhibit similar behavior with the advantage of the former being less computationally expensive \cite{brun2007}. However, in the case of complex models and clusters, an approach based on the minimization of prediction error may be particularly advantageous because internal indices tend to fail to correlate well with errors \cite{brun2007}.

\section*{Theoretical details} 
\label{sec:models}

\subsection*{Stability measure}
\label{sec:stab}
The notion of stability by Lange et al. \cite{lange2004} is used to assess solutions of clustering algorithms based on the rationale that true clusters are those that can always be identified by a clustering algorithm when applied to different datasets from the same generating process. Formally, let $\mathcal{A}_k$ be a clustering algorithm with $k$ the number of clusters, $\phi$ a classifier, and $(\mathbf{X}, \mathbf{Y})$ the training set and clustering labels, i.e., $\mathcal{A}_k(\mathbf{X})=\mathbf{Y}$. After training $\phi$ on $(\mathbf{X}, \mathbf{Y})$, both the clustering algorithm and trained classifier are applied to a separate dataset $\mathbf{X}'$. The distance between the two solutions is the normalized Hamming distance:
\begin{equation}
    d_{S_k}(\phi(\mathbf{X}'),\mathbf{Y}') = \min_{\sigma \in S_k}\frac{1}{n}\sum_{i=1}^{n}\mathbb{1}_{\{\sigma(\phi(X_i'))\neq Y_i'\}}
    \label{eq:diss}
\end{equation}
with $S_k$ the set of all possible permutations of $k$ elements. Supervised labels are permuted to overcome the non-uniqueness of clustering labeling and $\sigma$ is the permutation that minimizes the solutions dissimilarity. Averaging out the distance between any pair of partitions $\mathbf{X},\ \mathbf{X}'$ from Eq.~\ref{eq:diss} we define the stability index of the clustering algorithm as:
\begin{equation}
    \mathcal{S}(\mathcal{A}_k) = \E_{\mathbf{X},\mathbf{X'}}[d_{S_k}(\phi(\mathbf{X}'), \mathbf{Y}')]
    \label{eq:2}
\end{equation}  
The stability index ranges in $[0, 1]$, with lower values indicating more stable and reproducible solutions \cite{lange2004}. Because this measure scales with the number of clusters, the measure suggested by the authors is the normalized stability $\bar{\mathcal{S}}_k$, i.e., the stability from Eq.~\ref{eq:2} normalized for the stability of random labeling $\mathcal{R}_k$.

\subsection*{\pkg{reval} algorithm}
\begin{figure}[h!]
\centering
\includegraphics[width=\textwidth,keepaspectratio]{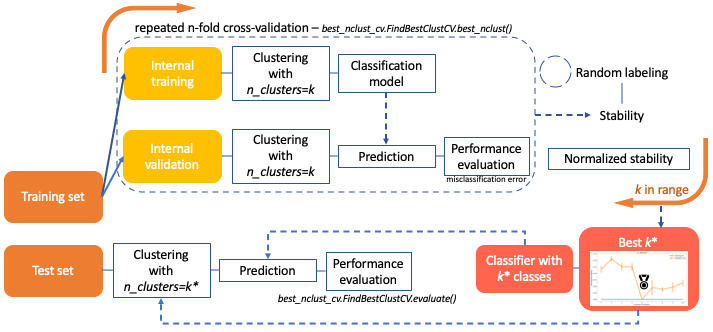}
\caption{\pkg{reval} implementation overview.
Repeated cross validation procedure is included within the dashed circle and it is repeated for different numbers of clusters $k$ as indicated by the orange arrows (Algorithm~\ref{alg1}). The clustering algorithm with number of clusters $k^*$, i.e., the maximum value that minimizes normalized stability, is evaluated on a held-out dataset (Algorithm~\ref{alg2}).}
\label{fig:pipeline} 
\end{figure}

The algorithm implemented in \pkg{reval} allows the user to: 1) automatically select the number of clusters for a dataset by minimizing validation stability, via repeated cross validation (see Algorithm~\ref{alg1}); and 2) compute classification performance obtained when generalizing the solution to an held-out dataset (see Algorithm~\ref{alg2}). An overview of the framework is reported in Figure~\ref{fig:pipeline}.

A dataset $\mathbf{X}$ is first split  into training $\mathbf{X}_{tr}$ and test $\mathbf{X}_{ts}$ sets and a clustering $\mathcal{A}$ and classifier $\phi$ are selected. Let $n_{\mathrm{fold}}$ be the number of folds for cross validation, $n_{r}$ the number of repetitions, $n_{rnd}$ the number of random labeling iterations, and $k$ the number of clusters in set $K$. In Algorithm~\ref{alg1} we indicate with $\mathbf{X}^{i_j}_{itr}$ and $\mathbf{X}^{i_j}_{val}$ the internal training and validation splits of training set $\mathbf{X}_{tr}$, respectively, for cross-validation $i$th fold split at the $j$th shuffled repetition. These correspond to $\mathbf{X}$ and $\mathbf{X'}$ sets introduced in the ``Stability measure'' section. With $(K\times n_{\mathrm{fold}} \times n_{r})$ we indicate the Cartesian product of the sets of number of clusters and repeated cross-validation splits. The fitted model becomes the one trained on $\mathbf{X}_{itr}$ that returns the maximum number of clusters with minimum stability. That model can then be used within Algorithm~\ref{alg2} for generalization on the test set.

Among clustering methods that work within \pkg{reval}, density-based clustering HDBSCAN \cite{campello2013} does not need any assumption on the number of clusters. Hence, we do not need to iterate over different number of clusters to select the best solution. Instead, normalized stability is computed within the repeated cross-validation loops that return the same number of clusters.

\vspace{5mm}

\begin{algorithm}[H]
\SetAlgoLined
\KwIn{$\mathbf{X}_{tr}$, $\mathcal{A}$, $\phi$, $K$, $n_{\mathrm{fold}}$, $n_{r}$, $n_{rnd}$}
\KwResult{$k^*$}
\For{$(k, i, j) \in (K\times n_{\mathrm{fold}} \times n_{r})$}{
Find clustering solution $\mathcal{A}_k(\mathbf{X}^{i_j}_{itr})=\mathbf{Y}^{i_j}_{itr}$ and train $\phi$ on ($\mathbf{X}^{i_j}_{itr}, \mathbf{Y}^{i_j}_{itr}$)\;
Compute $\phi_{i_j}(\mathbf{X}^{i_j}_{val})$ and $\mathcal{A}_k(\mathbf{X}^{i_j}_{val})=\mathbf{Y}^{i_j}_{val}$\; Select permutation $\bar{\sigma}_{i_j} \in S_k$ that yields to minimum dissimilarity $d_{\bar{\sigma}_{i_j}}(\phi_{i_j}(\mathbf{X}^{i_j}_{val}), \mathbf{Y}^{i_j}_{val})$\;
\For{$r = 1, \dots, n_{rnd}$}{
Train $\phi$ on ($\mathbf{X}^{i_j}_{itr}, \mathcal{R}_k(\mathbf{Y}^{i_j}_{itr})$)\; 
Compute $d_{\bar{\sigma}_{r_j}}(\phi_{r_j}(\mathbf{X}^{i_j}_{val}), \mathbf{Y}^{i_j}_{val}))$ as before\;}
Compute $d_{i_j}=d_{\bar{\sigma}_{i_j}}/\Avg_{r=1}^{n_{rnd}}(d_{\bar{\sigma}_{r_j}})$\;
Compute normalized stability $\bar{\mathcal{S}}_k=\Avg_{j=1}^{n_{r}}\Avg_{i=1}^{n_{\mathrm{fold}}} d_{i_j}$\;
}
Return $k^*$ s.t. $\max\argmin_{k\in K}\bar{\mathcal{S}}_k$.
\caption{Return number of clusters that minimizes normalized stability.}
\label{alg1}
\end{algorithm}

\newpage
\begin{algorithm}[H]
\SetAlgoLined
\KwIn{$\mathbf{X}_{tr}$, $\mathbf{X}_{ts}$, $k^*$, $\mathcal{A}_{k*}$, $\phi$}
\KwResult{classification accuracy}
Find clustering solution $\mathcal{A}_{k^*}(\mathbf{X}_{tr})=\mathbf{Y}_{tr}$ and train $\phi$ on ($\mathbf{X}_{tr}, \mathbf{Y}_{tr}$)\;
Compute $\phi_{k^*}(\mathbf{X}_{ts})$ and $\mathcal{A}_{k^*}(\mathbf{X}_{ts})=\mathbf{Y}_{ts}$\;
Compute the accuracy (ACC) with permuted clustering labels for consistency between training and test sets:
\begin{displaymath}
\ACC=\max_{\sigma \in S_{k^*}}\Avg_{i=1}^{|\mathbf{X}_{ts}|}\mathbb{1}_{\{\phi((X_{ts})_i)= \sigma((Y_{ts})_i)\}}
\end{displaymath}
Return $\ACC \in [0, 1]$.
\caption{Test best solution on unseen data.}
\label{alg2}
\end{algorithm}

\section*{Technical details}
\subsection*{Implementation}
The \pkg{reval} package has four core modules:

\begin{itemize}
\item \texttt{relative\_validation}: This module includes training and test methods that return misclassification errors obtained by comparing classification labels and clustering labels. It also includes the random labeling method which allows users to compute the asymptotic misclassification rate.

\item\texttt{best\_nclust\_cv}: This module implements repeated cross validation and returns the best clustering solution together with normalized stability scores, obtained from the average of the misclassification scores divided by the asymptotic misclassification rate. Repeated cross validation leads to unbiased stability estimates and it can also be performed stratifying the repeated randomized splits according to a desired variable. To control for the size imbalance that derives from cross validation we initialized the repeated cross-validation loop to a $1\times 2$ schema as default. Users can change this configuration according to dataset size and available stratifiers that can be useful to overcome imbalance issues. The evaluation method applies the fitted model with the returned number of clusters to the held-out dataset and returns accuracy (ACC). Other metrics, such as Matthews correlation coefficient (MCC) \cite{matthews1975}, F1 score, precision and recall scores can also be computed (see \emph{utils} file).

\item \texttt{param\_selection}: This module enables hyperparameter tuning to select the best configuration of classifier/clustering (\texttt{SCParamSelection} class) and the parameters within clustering and classifier themselves (\texttt{ParamSelection} class). Best parameters are those that report minimum normalized stability. If the number of true classes is available this module also returns the best solution that correctly identifies the true number of clusters, if it exists.

\item \texttt{visualization}: This module includes the function to plot cross-validation performance metrics with 95\% confidence intervals for a varying number of clustering solutions. The threshold of random labeling stability can be displayed to visually investigate model performance.
\end{itemize}

\begin{figure}[h]
    \centering
    \begin{subfigure}[t]{0.32\textwidth}
        \centering
        \includegraphics[width=\linewidth]{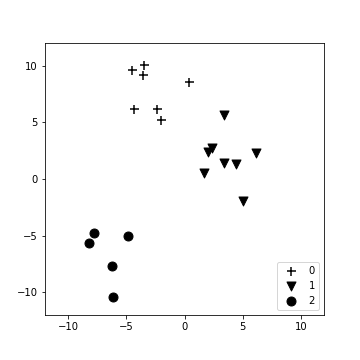} 
        \caption{} \label{fig:labelstr}
    \end{subfigure}
    \begin{subfigure}[t]{0.32\textwidth}
        \centering
        \includegraphics[width=\linewidth]{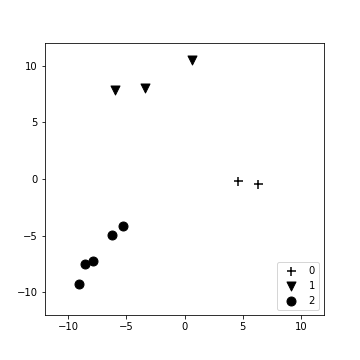} 
        \caption{} \label{fig:labelsts}
    \end{subfigure}
    \begin{subfigure}[t]{0.32\textwidth}
        \centering
        \includegraphics[width=\linewidth]{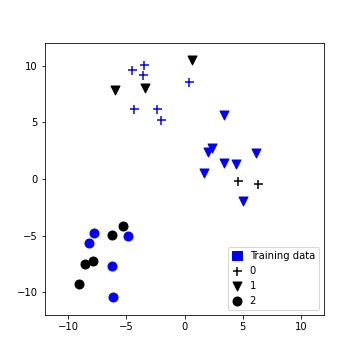} 
        \caption{} \label{fig:labelsall}
    \end{subfigure}
    \caption{Relabeling practice. Clustering labels on the training (a); and test (b) sets. Differences in labeling between training and test sets are displayed in (c). In blue the labeled points on which the classifier is trained, in black the labeled points whose classes we want to predict.}
    \label{fig:labels}
\end{figure}

As suggested by Lange and colleagues \cite{lange2004}, we used the Kuhn-Munkres algorithm \cite{kuhn1955, munkres1957} to obtain the label permutation that minimizes misclassification error. However, differently from the work of Lange and colleagues \cite{lange2004}, \pkg{reval} permutes the clustering labels instead of the classification labels, i.e., the normalized Hamming distance in Eq.~\ref{eq:diss} becomes:
\begin{displaymath}
    d_{S_k}(\phi(\mathbf{X}'),\mathbf{Y}') = \min_{\sigma \in S_k}\frac{1}{n}\sum_{i=1}^{n}\mathbb{1}_{\{\phi(X_i')\neq \sigma(Y_i')\}}
\end{displaymath}
This approach allows the test set to preserve the partition structure of the training set, to better investigate results replicability and to aid visual comparison. Figure~\ref{fig:labels} shows the rationale behind the need to permute clustering labels when training a classifier within \pkg{reval}. We simulated three Gaussian blobs and divided them into training ($N=20$) and test ($N=10$) sets. Figure~\ref{fig:labelstr} shows clustering labels for the training set and Figure~\ref{fig:labelsts} the clustering labels for the test set. Figure~\ref{fig:labelsall} shows what might happen when training a classifier on the training set labels and then predict labels for the test set. Because two out of three classes show label discordance the trained classifier fails to correctly predict the classes. Nevertheless, if we permute class labels $0$ and $1$ in the example, the trained classifier will correctly identified all three classes on the test set returning minimum prediction error and consistent label ordering is preserved. The \fct{kuhn\_munkres\_algorithm} function can be found in \emph{utils} file.

A more thorough description of the code and its usage can be found at \url{https://reval.readthedocs.io/en/latest/code_description.html}. \pkg{reval} mainly works with the \pkg{scikit-learn} \proglang{Python} library for machine learning \cite{sklearn}. In particular, among clustering methods, users can select those with number of clusters parameter, i.e., k-means, hierarchical clustering, and spectral clustering, but also density-based clustering HDBSCAN \cite{campello2013} from \pkg{hdbscan} library. Moreover, any classifier from \pkg{scikit-learn} can be selected.

\subsection*{Algorithm complexity}
We report here the complexity analysis of the two core methods included in \texttt{best\_nclust\_cv} module. In particular, we focus on \fct{best\_nclustcv}, which enables the selection of the best number of clusters via repeated cross validation, and \fct{evaluate}, which implements the testing of the best solution on held-out datasets. The \fct{best\_nclustcv} method includes sequential calls to train (\fct{train}), test (\fct{test}), and random labeling (\fct{rndlabels\_traineval}) methods from the \texttt{relative\_validation} module, whereas \fct{evaluate} sequentially calls the train and test functions.

The overall complexity to perform cross validation and evaluation depends on the number of calls to the relative validation module functions and their intrinsic cost. In particular, the complexity of the training method is led by the sum of the costs of the chosen classification ($O(\text{classifier})$) and clustering ($O(\text{clustering})$) algorithms, which usually depend on the number of data samples and features. The complexity of the \fct{test} method mainly depends on the complexity of the clustering algorithm in that only prediction is performed for the classifier. The random labeling algorithm increases the classifier complexity by a factor of $n_{rnd}$ (i.e. the number of random labeling iterations). To perform cross validation and compute normalized stability we need to set the following parameters: $n_{C} = |K|$ (i.e. the length of the list with the different number of clusters to try); $n_{\mathrm{fold}}$ and $n_{r}$, which corresponds to the number of cross-validation folds and repetitions respectively; and $n_{rnd}$. In conclusion, the cost of \fct{best\_nclustcv} is equal to $n_{C} \cdot n_{\mathrm{fold}} \cdot n_{r} (O(\text{classifier}) + O(\text{clustering}) + n_{rnd} O(\text{classifier}))$ and the complexity of \fct{evaluate} is $O(\text{classifier}) + 2 \cdot O(\text{clustering})$.

\begin{figure}[h]
    \centering
    \begin{subfigure}[t]{0.32\textwidth}
        \centering
        \includegraphics[width=\linewidth]{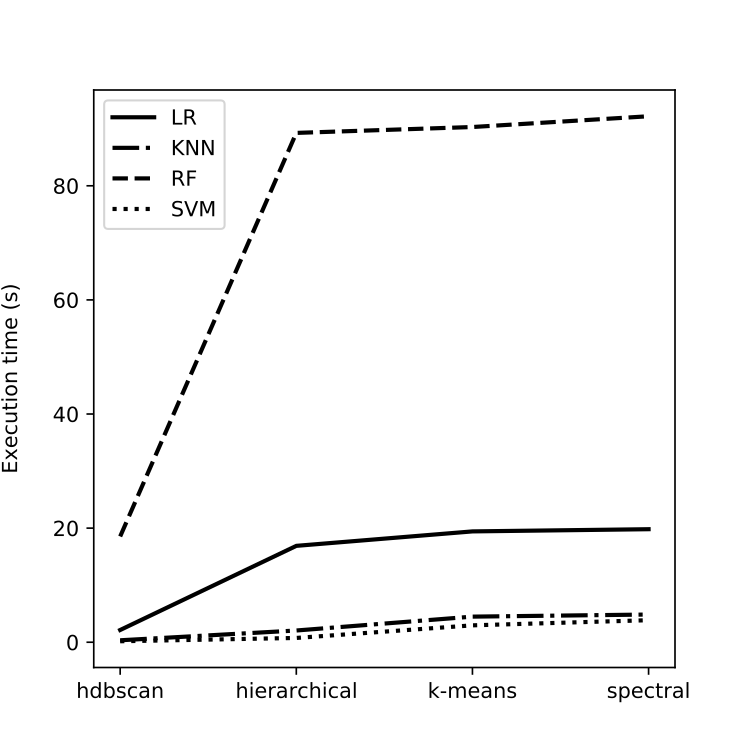} 
        \caption{} \label{fig:cv1}
    \end{subfigure}
    \begin{subfigure}[t]{0.32\textwidth}
        \centering
        \includegraphics[width=\linewidth]{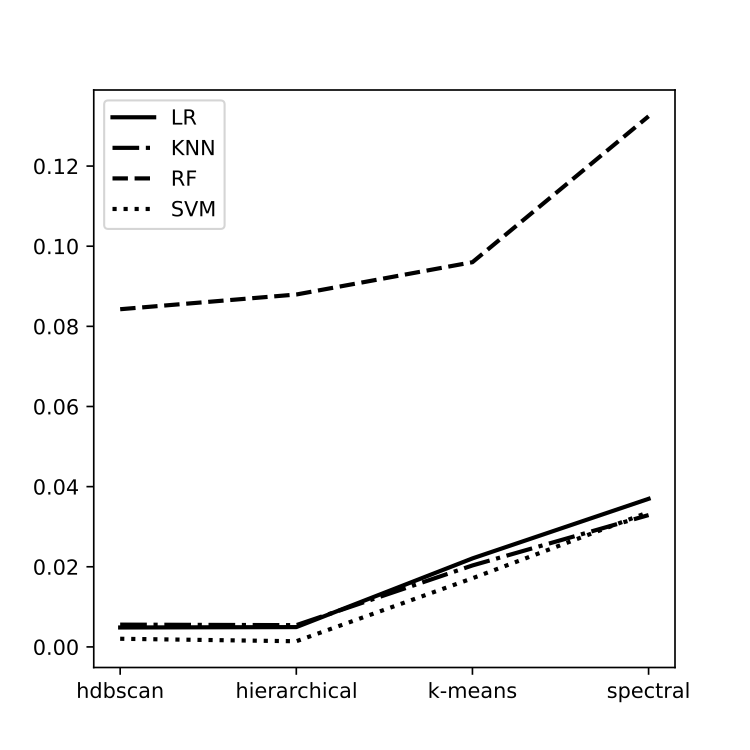} 
        \caption{} \label{fig:ev1}
    \end{subfigure}
    \begin{subfigure}[t]{0.32\textwidth}
        \centering
        \includegraphics[width=\linewidth]{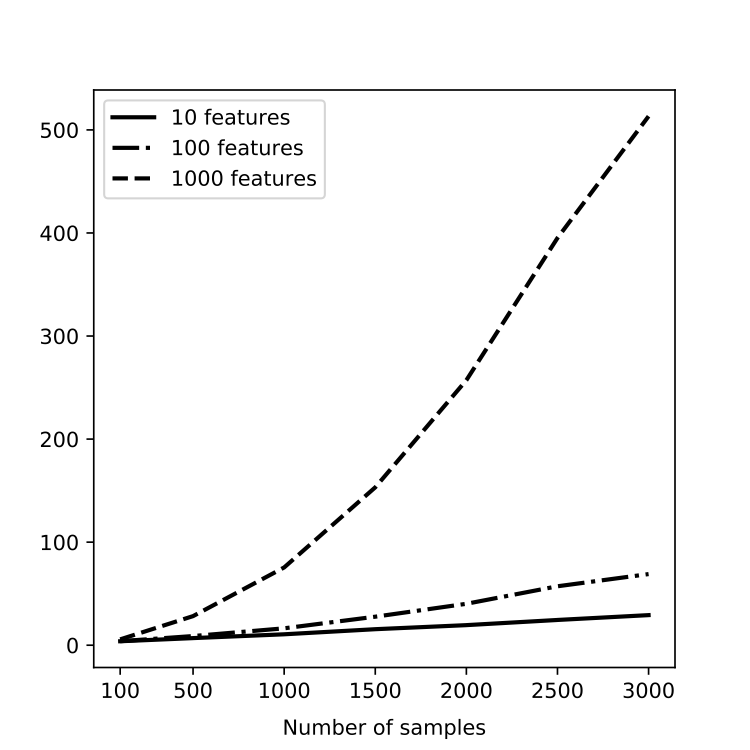} 
        \caption{} \label{fig:knnkm1}
    \end{subfigure}
    \caption{Execution times. Different combinations of classification and clustering algorithms applied to blobs dataset with $100$ samples and $10$ features. \fct{best\_nclustcv} (a); \fct{evaluate} (b); sequential calls to \fct{best\_nclustcv} and \fct{evaluate} (c) with k-nearest neighbors (KNN) and k-means for blobs dataset with varying combinations of samples and features.}
\end{figure}

We run the methods considered on simulated datasets to empirically investigate the execution times. Simulations were run on a MacBook Pro 2020 with a 2.3 GHz Quad-Core Intel Core i7 processor and 32 GB of RAM. The complexity of clustering algorithms and state-of-the-art classifiers that can be used with \pkg{reval} can be found in Table~\ref{tab:cmplx}. Figure~\ref{fig:cv1} and \ref{fig:ev1} show the execution times of \fct{best\_nclustcv} and \fct{evaluate}, respectively, on a two-blob dataset of $100$ samples and $10$ features with different combinations of classifiers and clustering algorithms. $n_{C}$ is set at $5$, $n_{\mathrm{fold}}$ and $n_{r}$ are $2$ and $10$, respectively, and $n_{rnd}$ is equal to $10$. Default parameters were used for all classifiers and clustering algorithms. It is straightforward to observe that execution times largely depend on the chosen algorithms, with HDBSCAN the least expensive and spectral clustering the most expensive choice among clustering techniques, irrespective of classifier, and random forest the most expensive choice among classifiers. Overall the execution time of the \fct{evaluate} method is reduced when compared to repeated cross validation. The package implementation also includes a multiprocessing feature which speeds up computations for repeated cross validation but not for the evaluation method, see Supplementary Figure~\ref{fig:cv7}, where $7$ jobs are simultaneously run. Figure~\ref{fig:knnkm1} shows the execution times with varying number of samples and features when sequentially running \fct{best\_nclustcv} and \fct{evaluate} with k-nearest neighbors (KNN) classifier and k-means clustering. We can observe a moderate increase for low number of samples ($<10^3$) with varying number of features and a steep growth for $10^3$ samples. Corresponding execution times with multiprocessing can be observed in Supplementary Figure~\ref{fig:knnkm7}. 

\section*{Technical validation} 
\label{sec:simulations}
To investigate \pkg{reval} performance, first we investigate group detection in simulated non-overlapping Gaussian blobs, and the handwritten digits dataset from the MNIST digit recognition database (Modified National Institute of Standards and Technology database - \url{http://yann.lecun.com/exdb/mnist/}). Then, we leverage the \texttt{SCParamSelection} class to determine the combination of classifier and clustering algorithms that best identifies the number of classes for $19$ different datasets from the UCI Machine Learning repository \cite{dua2019}.

\subsection*{Blobs dataset}
To provide a simple example of how to use \pkg{reval}, we generated $5$ isotropic Gaussian blobs for a total of $1000$ samples with $2$ features. To run the example refer to the code in the ``Blobs dataset'' section of the supplementary material. The dataset was first split into training and test sets ($70/30\%$). Then we selected KNN with the number of neighbors equal to $15$ and k-means with Euclidean distance, as suggested by the experiment reported in the ``Algorithm selection'' section. We then run the stability-based algorithm with a $10\times 2$ repeated cross validation framework, with $10$ random labeling repetitions, and the number of clusters varying from $2$ to $6$. We evaluated the solution found on the test set and report ACC and MCC scores as metrics. Last, we report the AMI score to compare true labels and predicted labels on the test set. We also determined the best number of clusters maximizing and minimizing internal measures such as silhouette and Davies-Bouldin. We computed the metrics independently on the training and test sets and return the best number of clusters found with k-means. 

\begin{figure}[h!]
\centering
\includegraphics[width=12cm,keepaspectratio]{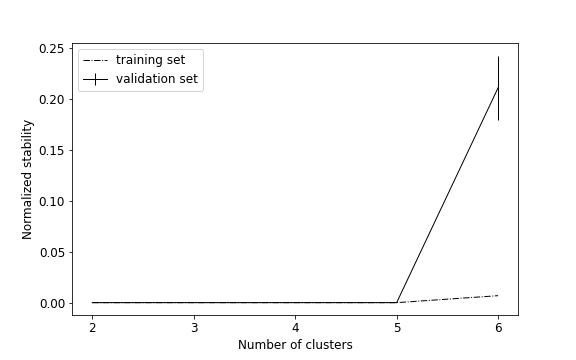}
\caption{\pkg{reval} performance for blobs dataset. Solid line represents the validation normalized stability with $95\%$ confidence intervals. Dashed line shows stability during training.}
\label{fig:blobsmetrics} 
\end{figure}

The normalized stability for varying number of clusters in Figure~\ref{fig:blobsmetrics} shows that the model perfectly identified $5$ clusters with $0.0$ normalized stability. The comparison between true and clustering labels can be observed in Supplementary Figure~\ref{fig:blobstest1}. The ACC and MCC scores on the test set are equal to $1.0$ and the AMI is equal to $1.0$. The maximum silhouette score is $0.83$ on both training and test sets with the correct number of clusters and AMI scores equal to $1.0$. The Davies-Bouldin index results do not replicate between training and test sets. The index is equal to $0.23$ with $4$ clusters on the training set and to $0.23$ with $5$ clusters on the test set. AMI scores are $0.91$ for training and $1.0$ for testing.

\subsection*{MNIST dataset}
For the real-world dataset example, we considered the handwritten digits MNIST dataset (\url{http://yann.lecun.com/exdb/mnist/}) that includes $70,000$ samples corresponding to $28\times 28$ images of digits from $0$ to $9$. When flattened, each sample has $784$ features. The dataset has $10$ classes, with $\sim7000$ samples for each class.

First, we split the dataset into training and test sets of $60,000$ and $10,000$ samples, respectively. Then, we preprocessed the dataset with UMAP \cite{mcinnes2018} for dimensionality reduction with $2$ components. We run HDBSCAN clustering (as suggested in \url{https://umap-learn.readthedocs.io/en/latest/clustering.html}) with different classifiers and selected the best configuration. In particular, we considered KNN with the number of neighbors equal to $30$, and SVM, RF, LR with default parameters from \pkg{scikit-learn} library. The HDBSCAN algorithm was initialized with minimum samples equal to $10$ and minimum cluster size equal to $200$. The relative clustering validation procedure was run with $10$ repetitions of $2$-fold cross validation, and number of random labeling iterations equal to $10$. Because HDBSCAN does not need the number of clusters specified a priori, the normalized stability is computed averaging the misclassification error over the solutions that return the same number of clusters. We selected the one that minimizes the normalized stability and the trained classifier applied to test set is the one trained on the corresponding cross-validation fold. Because we are not preselecting the number of clusters results might differ between training and test sets.

\begin{figure}[h!]
\centering
\includegraphics[width=12cm,keepaspectratio]{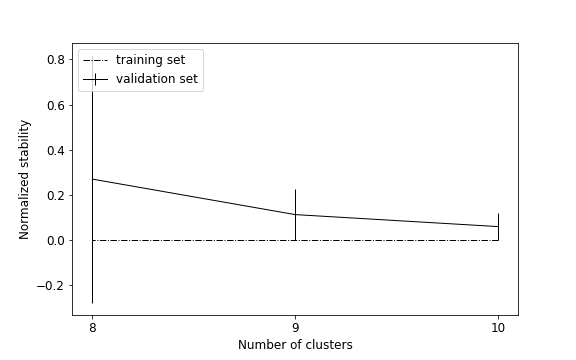}
\caption{\pkg{reval} performance for MNIST dataset with RF and HDBSCAN algorithms. Solid line represents the validation normalized stability with $95\%$ confidence intervals. Dashed line shows training stability.}
\label{fig:mnistmetrics} 
\end{figure}

\begin{figure}[h]
    \centering
    \begin{subfigure}[t]{0.475\textwidth}
        \centering
        \includegraphics[width=\linewidth]{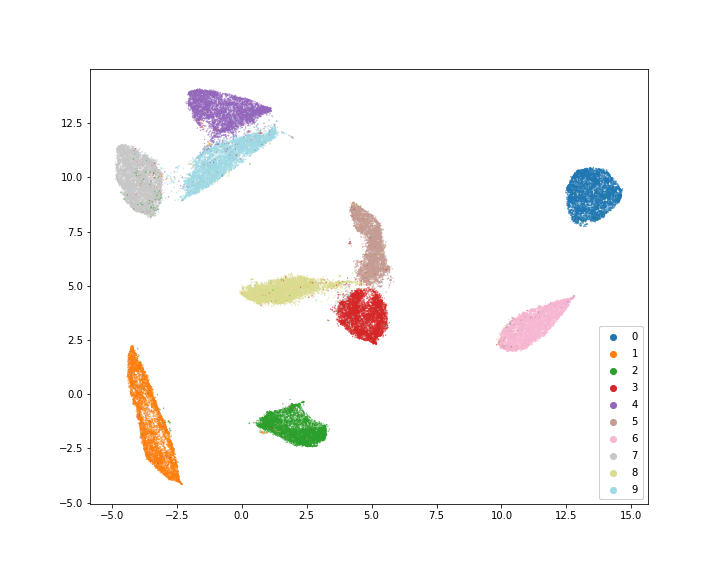} 
        \caption{}
        \label{fig:mnist}
    \end{subfigure}
    \begin{subfigure}[t]{0.475\textwidth}
        \centering
        \includegraphics[width=\linewidth]{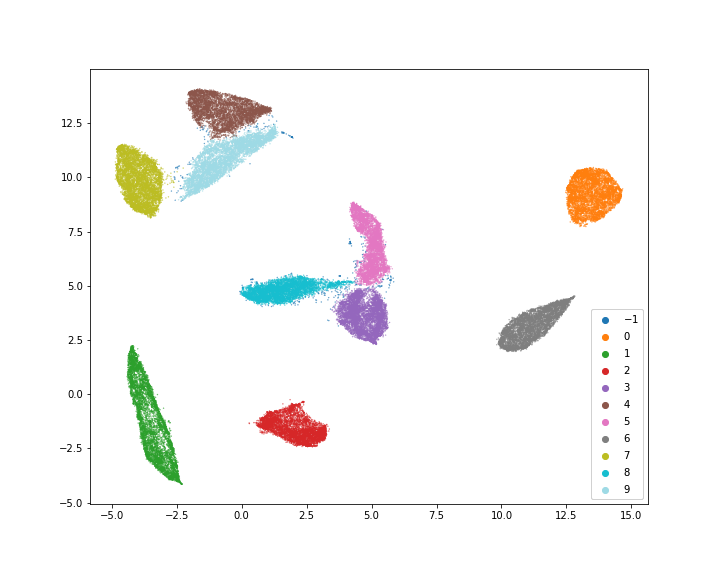} 
        \caption{}
        \label{fig:mnistreval}
    \end{subfigure}
    \vskip\baselineskip
    \begin{subfigure}[t]{0.475\textwidth}
        \centering
        \includegraphics[width=\linewidth]{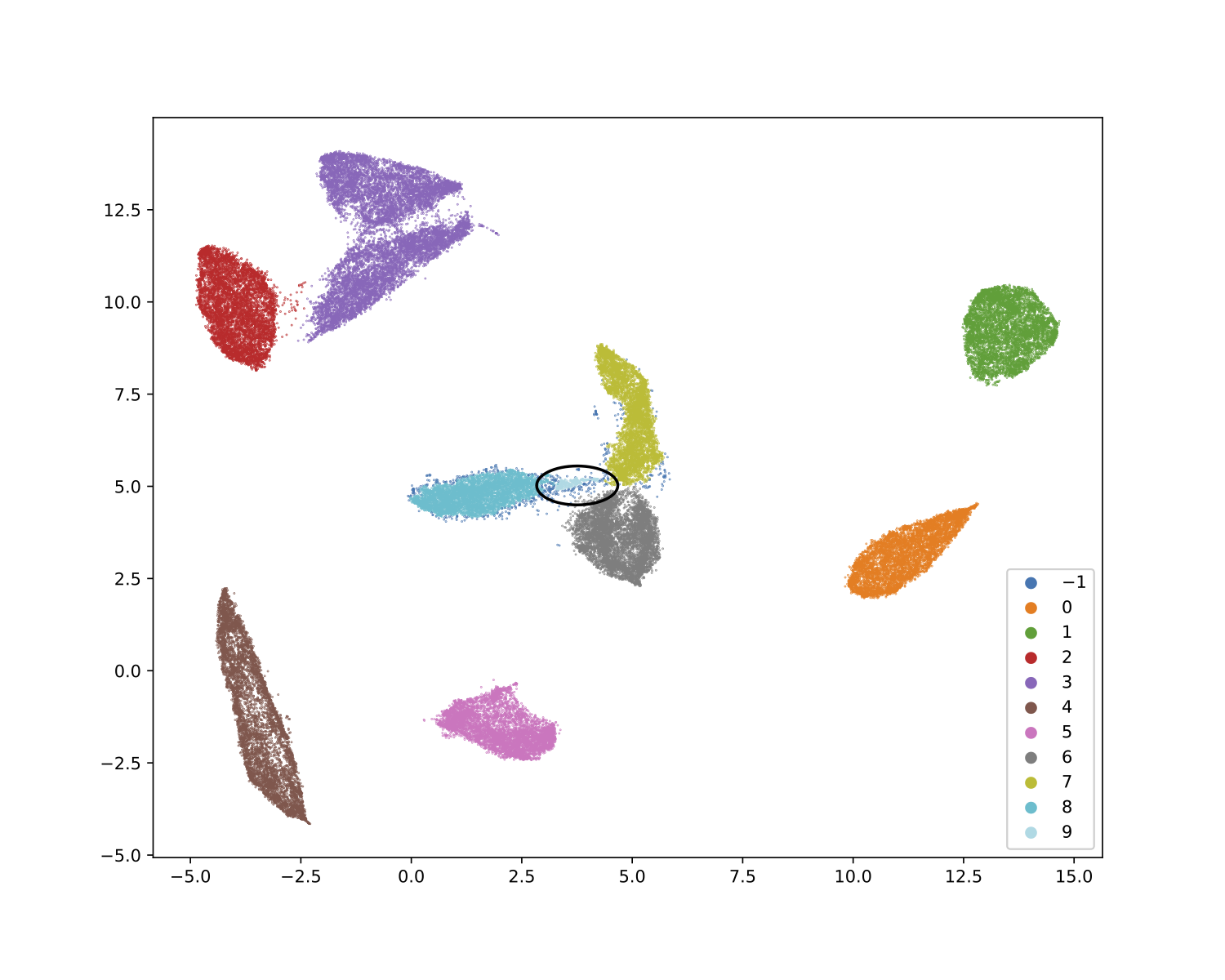} 
        \caption{} 
        \label{fig:mnistsilh}
    \end{subfigure}
    \begin{subfigure}[t]{0.475\textwidth}
        \centering
        \includegraphics[width=\linewidth]{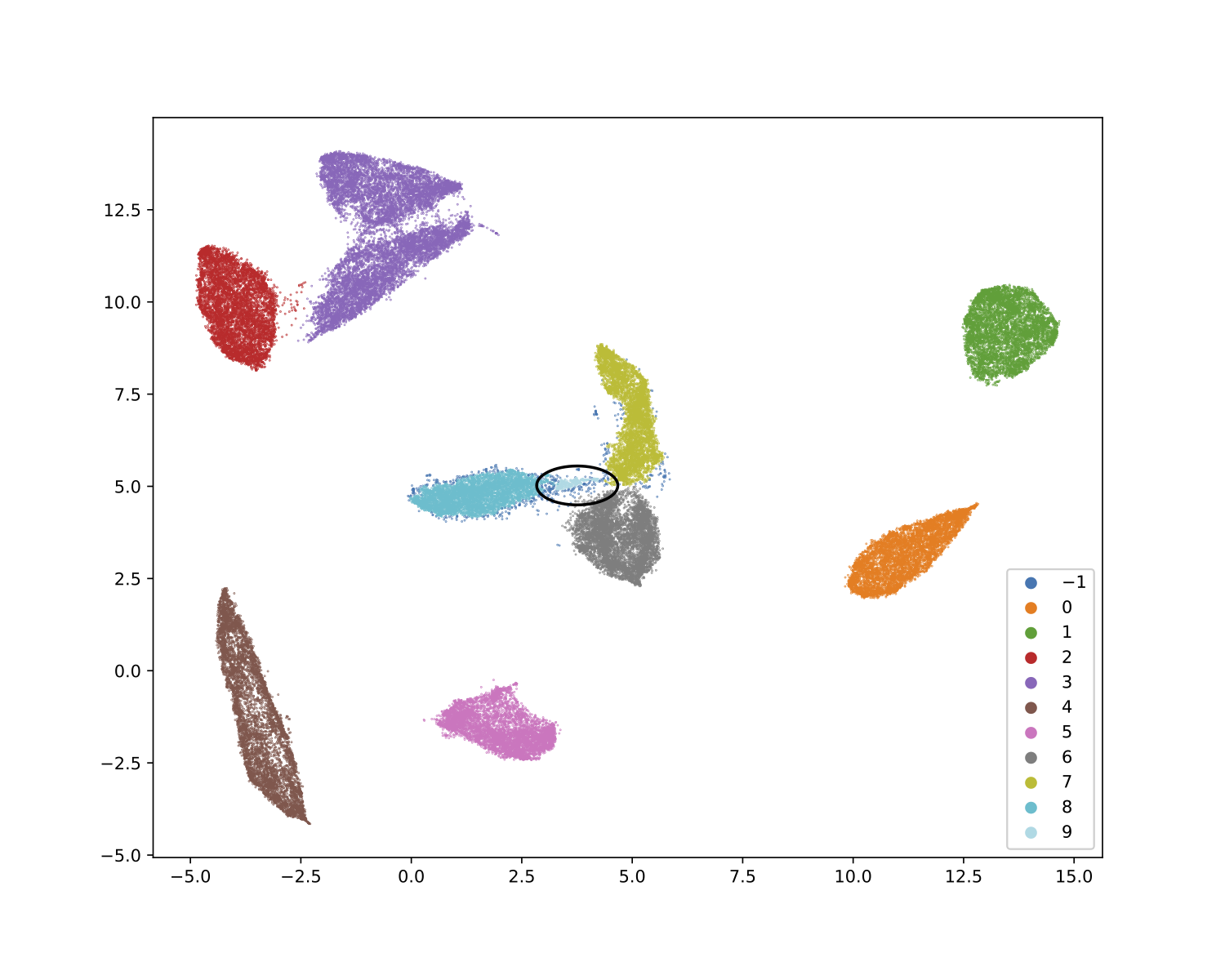} 
        \caption{} 
        \label{fig:mnistdb}
    \end{subfigure}
    \caption{MNIST cluster visualization. Training set uniform manifold approximation and projection (UMAP) visualization for true labels (a); relative validation labels (b); silhouette score labels (c); Davies-Bouldin index labels (d). For internal measures we circled the erroneous cluster identified.}
    \label{fig:intcomp}
\end{figure}

The best classifier-clustering combination is RF and HDBSCAN, which correctly identifies $10$ clusters with a misclassification error equal to $0.06\ (0.06)$ on the training set (see Figure~\ref{fig:mnistmetrics}), and replicates on the test set, identifying $9$ clusters with ACC equal to $0.86$ and MCC equal to $0.86$. The AMI score on the test set is equal to $0.87$. When the experiment was run with internal validation measures, we obtained $10$ clusters in training and $9$ clusters during testing, with silhouette scores equal to $0.60$ and $0.64$, respectively, and Davies-Bouldin scores equal to $0.96$ and $0.82$, respectively. In Figure~\ref{fig:intcomp}, we observe that internal measures fail to detect the actual digit classes during training, whereas \pkg{reval} with HDBSCAN and RF successfully identifies them (see Figures~\ref{fig:mnistsilh}, \ref{fig:mnistdb}). On test set the result is the same among all methods as demonstrated by equal AMI scores, see Supplementary Figure~\ref{fig:mnisttest}. Based on these results, we can speculate that clustering on a subset of data ($N=35,000$) better detects the digits classes.

It is worth noting that the classifier performance highly depends on the clustering solution in that it tends to overfit to the training dataset. To guide the choice of a classifier, users should first select an appropriate clustering algorithm, then select the classifier that minimizes stability and that reports the best performance on the held-out dataset. In case of equal or very similar outcomes, algorithm complexity should guide the classifier selection.

\subsection*{Algorithm selection}
We considered $19$ datasets from the UCI Machine Learning repository \cite{dua2019} including the test set of the handwritten digits dataset that can be found in \pkg{scikit-learn} ``toy datasets'' - \url{https://scikit-learn.org/stable/datasets/index.html#toy-datasets}. In Table~\ref{tab:dataset} we report the dataset names along with the number of samples, features, and inherent classes. We applied the relative validation algorithm with different combinations of classifier and clustering algorithms. In particular, for clustering we selected: hierarchical clustering (HC) with Ward's method and Euclidean distance; k-means clustering with Euclidean distance; and HDBSCAN with minimum cluster size equal $5$ and Euclidean distance. Among classifiers we opted for: KNN with $1$ neighbor and Euclidean distance; random forest (RF) classifier with $100$ estimators; support vector machines (SVM) with $C=1$ and $\gamma=\frac{1}{\text{N samples}}$; and logistic regression (LR). We did not consider spectral clustering because of its computational cost (see Table~\ref{tab:cmplx}). To improve the performance, in addition to raw datasets, we repeatedly run the experiments after preprocessing with: 1) standard scaler, which removes the mean and scales to unit variance; 2) uniform manifold approximation and projection (UMAP) algorithm \cite{mcinnes2018} for dimensionality reduction; and 3) standard scaler and UMAP. UMAP parameters are chosen according to those suggested in the documentation - \url{https://umap-learn.readthedocs.io/en/latest/clustering.html}. We ran the algorithm with $10$ replications of 2-fold cross validation and $10$ random labeling iterations. The number of clusters ranges from $2$ to $n_{\mathrm(classes)}+2$, where $n_{\mathrm{classes}}$ is the number of classes for each dataset.

Table~\ref{tab:result} reports the best solutions selected as those reporting minimum stability along with the correct number of clusters. If no experiment identified the correct number of classes, we chose, among the solutions with minimum stability for each preprocessed dataset, the one with maximum adjusted mutual information (AMI), ACC, and MCC. AMI external score measures the similarity of two labelings of the same data, irrespective of label order and ranges from $[0, 1]$, with a perfect match equal to $1$. For comparison, we computed the silhouette score and Davies-Bouldin index \cite{davies1979} internal measures independently on both training and test sets with the clustering algorithm selected by the stability-based approach and with the same varying number of clusters. We report in Table~\ref{tab:resultint} the best solutions as those that maximize and minimize those measures respectively. Davies-Bouldin index \cite{davies1979} measures clusters separation and ``tightness'' with Euclidean distance. Lower indices correspond to better solutions. AMI scores are reported to evaluate the similarity of true and cluster labels on test sets.

We observe that the stability-based approach identified the correct number of classes in $68\%$ of cases ($N=13$). Moreover, $6$ out $13$ experiments selected k-means clustering and KNN classifier as the best choice. Of these, only $1$ utilized raw data with no preprocessing, whereas the rest required either UMAP-preprocessed and/or scaled datasets. Because k-means works well with center-based spherical clusters and usually cannot find a good representation if clusters are very elongated or have complicated shapes, data preprocessing can relax this issue. From this experiment, it emerged that UMAP can be used as a preprocessing tool in this sense, in that it unwraps manifolds to find manifold boundaries. Nevertheless, because each dataset has its own intrinsic characteristics, preprocessing steps must be chosen with care to avoid breaking clusters into several erroneous small spherical clusters (see example with k-means at \url{https://umap-learn.readthedocs.io/en/latest/clustering.html}). It has been observed \cite{luxburg} that if the number of clusters $k$ is too large with respect to the true clusters, the k-means algorithm tends to be unstable. On the contrary, if $k$ is smaller or equal to the true number of clusters, the algorithm tends to be stable. Therefore, it is argued that k-means stability depends on the number of true clusters in the dataset, which should be on the order of $10$ to provide stable solutions. This seems to hold when we look at the UCI datasets with $\geq 10$ classes for which k-means is often selected as the best clustering algorithm although it returns smaller number of clusters respect to true classes. For this reason, despite KNN/k-means provides the best algorithm configuration in more than half the experiments, the choice of a classifier/clustering should be done carefully taking into account the dataset dimension and the computational cost of the algorithms.

The number of clusters selected with the silhouette score and Davies-Bouldin index (see Table~\ref{tab:resultint}) is equal to that reported in Table~\ref{tab:result}. The exception here was the \emph{iris} dataset \cite{fisher1936}, whereby \pkg{reval} selects 3 clusters during training and 2 during testing with RF and HDBSCAN, whereas internal measures result in 4 and 2 cluster solutions. Nevertheless, the validation stability of the relative approach is equal to $0.33\ (0.19)$, suggesting that the partition does not generalize well because the solution is not stable. If we could only rely on internal measures we would have failed to acknowledge the quality of the solution found. Generally, because 2 out of 3 classes are not linearly separable and the data is displayed into two separate groups \cite{fisher1936} the \emph{iris} dataset is not a good candidate for clustering and only the relative validation approach can clearly show that.

More generally, when comparing the AMI scores for the internal measure results, we have no ability to say how well the results are actually doing. On the contrary, we have a sense of how good the clustering is with relative validation through the generalization process. See the example of the \emph{climate} dataset, which has a validation normalized stability of $0.20\ (0.04)$, indicating poor generalization, and a silhouette score on test set of $0.49$, although both methods provide the same clustering solution on test set given that AMI score is equal to $-0.005$ in both cases. 

In conclusion, grid-search for classifiers and clustering methods is easily and effectively implemented with \pkg{reval} and can be handy to avoid a priori selection of a classifier. Furthermore, the stability-based approach helps evaluate the goodness of a clustering solution by means of its generalization process. This information about generalization is absent with internal measures.

\section*{Stability regime to guide cluster selection}
Ben-Hur and colleagues presented a stability-based method with data sub-sampling and reported on the risk of underestimating/overestimating the true number of clusters \cite{ben2001}. Introducing prediction strength computed via repeated cross validation, Tibshirani et al. \cite{tibshirani2005} linked unsupervised to supervised learning in the attempt to overcome the best cluster estimation issue. The \pkg{reval} implementation moves forward adding the evaluation of the best solution on unseen data, an approach that is particularly suitable for datasets with large sample size. While large sample sizes were less common in the past, dataset size has increased substantially over time and \pkg{reval} can be particularly important and well suited for modern data science contexts. Solution generalizability can be leveraged to select the best number of clusters not only based on validation metrics (e.g., prediction strength greater than $0.8/0.9$ \cite{tibshirani2005}), but also on the performance of the solution applied to a new set of data. In this way, we are able to compare test set distribution to the one on which the result was based, hence reinforcing the decision. Selecting the number of clusters that best generalizes to new data (i.e., investigation of the stability regime) holds promise for overcoming the underestimation issue. To give a better sense of how this works in practice we present a case study in the context of autism research, where clustering solutions within a stability regime could be further investigated. 

Autism spectrum conditions (ASCs) are characterized by difficulties in social-communication alongside heightened restricted and repetitive behaviors \cite{dsm5}. The spectrum of affected individuals with ASC is highly heterogeneous and this heterogeneity is present at multiple levels, from genome to phenome, and can co-exist with differing levels of severity and comorbidities \cite{lai2014, lombardo2019}. Given the high level of heterogeneity, data-driven clustering could be a promising approach to isolating different types of autisms. To split ASC into data-driven subtypes, we applied \pkg{reval} to clinical data obtained from the National Database for Autism Research (NDAR - \url{https://nda.nih.gov/}). NDAR is a database that includes a heterogeneous collection of de-identified human subjects data for autism research. We focus here on clinical behavioral data from the Vineland Adaptive Behavior Scales (VABS) \cite{vinelandii2005, vineland3}. Within the VABS, there are three domain total scores for communication, living skills, and socialization skills. Using these 3 domain total scores with UMAP preprocessing, we trained the stability-based model on $420$ subjects (mean age $41.65\ (17.91)$ months, female/male counts $109/311$) . As for analysis choices, we ran this analysis with $100\times 3$ repeated cross validation and $100$ iterations of random labeling, with number of clusters ranging from $2$ to $10$, using k-means clustering, and a KNN classifier with number of neighbors equal to $15$. 

\begin{figure}[h!]
\centering
\includegraphics[width=12cm,keepaspectratio]{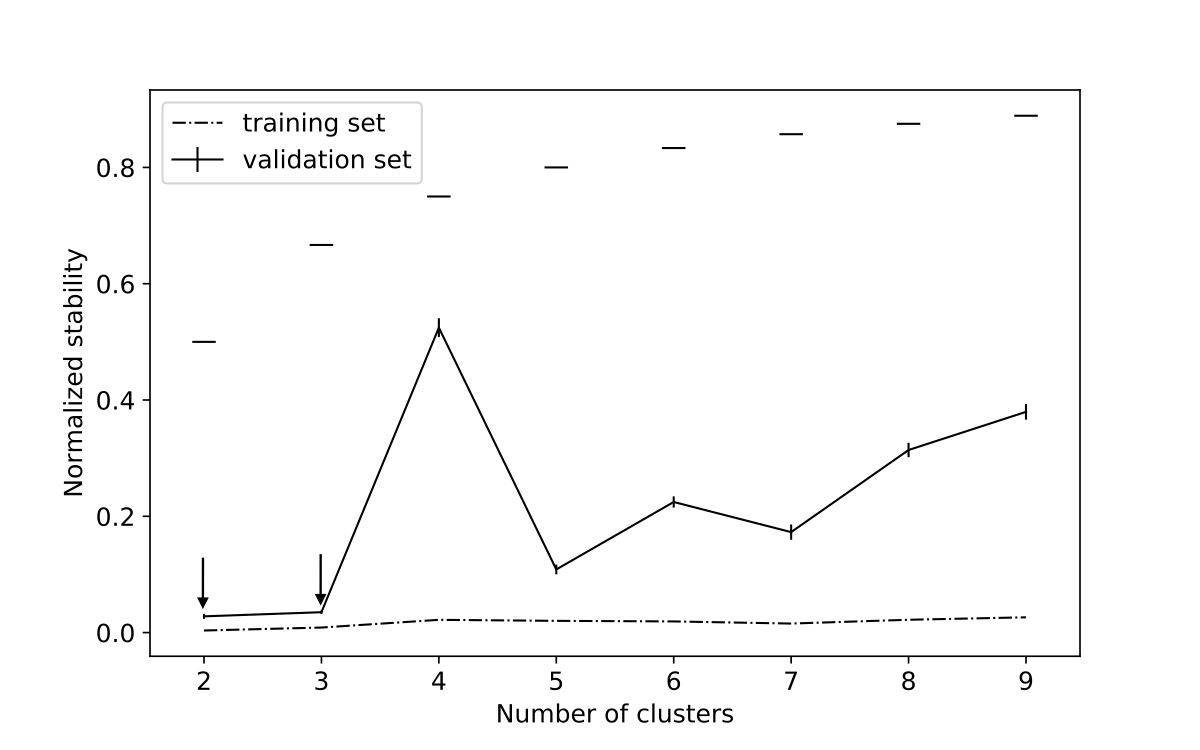}
\caption{Stability regime for the national database for autism research (NDAR) dataset. Random label stability is displayed for performance evaluation. Arrows point to the stability-regime solutions.}
\label{fig:ndarperf}
\end{figure}

From the performance plot in Figure~\ref{fig:ndarperf} we find that both a 2-cluster and 3-cluster solution results in small stabilities [2-cluster solution (error): $0.027\ (0.004)$; 3-cluster solution (error): $0.036\ (0.003)$] and thus form a stability regime whereby either solution might be a promising solution to follow-up with future work. Based on the minimization of the normalized stability measure, the default behavior when using \pkg{reval} would be to select $2$ as the best number of clusters. However, upon evaluation of these solutions on the unseen test set (n = $344$ subjects; mean age $43.12\ (17.04)$ months, female/male counts $90/254$), we find that both solutions reach $94\%$ accuracy, further confirming the presence of a stability regime whereby more than one solution might be a plausibly good model for follow-up work. The generalization performance alongside visual inspection (see Supplementary Figure~\ref{fig:ndar}) indicates that the default selection of a 2-cluster solution would possibly underestimate the true number of clusters \cite{tibshirani2005}. Thus, based on the stability regime present here, we could select $3$ as the true number of clusters since the stability differences are likely neglibgle and because both 2 and 3 cluster solutions generalize equally well. In practice, we would ultimately follow-up with examination of both 2 and 3 cluster solutions and utilize other datasets to better understand which solution might be most illuminating for decomposing clinical and biological heterogeneity of importance for autism research. 

To contrast the \pkg{reval} stability regime results here with internal measures, we would have obtained $2$ cluster solutions for both silhouette (scores of $0.41$ and $0.42$ on training and test respectively)  and Davies-Bouldin measures ($0.86$ on both training and test sets). If we force the number of clusters to $3$ we obtain lower silhouette scores (i.e., $0.35$ in both training and test sets), but higher Davies-Bouldin indices (i.e., $0.93$ and $0.95$, respectively). In this example, internal measures do not reveal a regime of possible cluster solutions. Given the additional lack of information about generalization from internal measures, such a regime might be easily missed. This example illustrates a real-world example in data science for how relative validation implemented with \pkg{reval} may reveal insights regarding regimes of clustering solutions that may be missed with internal validation approaches.

\section*{Discussion}
In this work, we introduce the \pkg{reval} package for relative clustering validation and describe how it can be utilized as well as provide examples for how it performs in simulations and several real datasets. In many cases \pkg{reval} successfully identifies the correct number of clusters and confers several other advantages over and above other internal validation approaches. First, when compared to internal validation measures, \pkg{reval} is able to report the extent different clustering solutions fit to the data at hand and how well those solutions may generalize or replicate on unseen data. Moreover, because \pkg{reval} works with multiple clustering algorithms, it can facilitate a more thorough investigation of clustering mechanisms. In fact, although a thorough theoretical and experimental analysis of stability-based model selection with k-means clustering have been done \cite{luxburg, ben2001, lange2004}, the effectiveness of such an approach needs to be further investigated with different clustering algorithms. Furthermore, \pkg{reval} can be included in ensemble learning pipelines \cite{rodriguez2018} or integrated in ensemble clustering frameworks for the selection of the best clustering solution \cite{strehl2002}. Last, the ability to identify stability regimes and evaluate such regimes based on generalization to unseen data can inform the underestimation issue of the best number of clusters identified in real-world datasets. 

A primary caveat or limitation to the approach \pkg{reval} takes is primarily one of data size. \pkg{reval} identifies the best clustering solution within a cross-validation framework, and hence needs large sample size to preserve cluster distribution between training and validation sets. Moreover, a separate held-out dataset is also needed to generalize the solution found. Smaller sized datasets may not allow for sufficient splitting within a cross-validation framework to allow for robust clustering solutions to generalize in unseen datasets. Finally, \pkg{reval} does not address the possibility of finding unrealistic data partitions. Because classifiers can overfit to their training set, a stable solution does not necessarily imply the true presence of subgroups in the data.  Future work will focus on the implementation of other relative validation methods, e.g., based on prediction strength \cite{tibshirani2005} with the aim to create a comprehensive library.

\section*{Computational details and code availability}
Code, package installation instructions, and documentation with working examples, can be found at \url{https://github.com/IIT-LAND/reval\_clustering}. The code to replicate the simulations presented in the ``Technical validation'' section is in \texttt{working\_examples.manuscript\_examples.py}. Code was written in Python 3.8 and simulations were run on a MacBook Pro 2020 with a 2.3 GHz Quad-Core Intel Core i7 processor and 32 GB RAM.

\clearpage
\section*{Acknowledgments}
This project was supported by funding from the European Research Council (ERC) under the European Union's Horizon 2020 research and innovation programme under grant agreement No 755816 (ERC Starting Grant to MVL).

\section*{Author contributions}
Conceptualization, I.L. and M.V.L; Methodology, I.L.; Software, I.L. and V.M.; Investigation,
I.L. and V.M.; Writing – Original Draft, I.L. and M.V.L.; Writing –
Review \& Editing, I.L., V.M., and M.V.L.; Funding Acquisition, M.V.L.; Supervision, M.V.L.

\section*{Declaration of interests}
The authors declare no competing interests.

\newpage
\singlespacing
\setlength\bibitemsep{8pt}
\printbibliography[title={References}]

\newpage
\begin{table}[t!]
\centering
\begin{tabular}{lcc}
\hline
Algorithm & Complexity & Problem\\ 
\hline
HDBSCAN & $O(N\log N)$ & clustering\\
K-means & $O(kNT)$ & clustering\\
Agglomerative & $O(kN^2)$ & clustering\\
Spectral & $O(N^3)$ & clustering\\
\hline
Logistic regression&$O(pN)$&classification\\
K-nearest neighbors&$O(pN)$&classification\\
Support vector machine&$[O(pN^2);O(pN^3)]$&classification\\
Random forest&$O(pN^2n_{trees})$&classification\\
\hline
\multicolumn{3}{l}{\footnotesize $N=\text{number of samples}$; $p=\text{number of features}$}\\
\multicolumn{3}{l}{\footnotesize K-means: $k=\text{number of clusters}$; $T=\text{number of iterations}$}\\ \multicolumn{3}{l}{\footnotesize Agglomerative: $k=\text{number of clusters}$}\\
\multicolumn{3}{l}{\footnotesize Random forest: $n_{tress}=\text{number of trees in the forest}$}
\end{tabular}
\caption{Algorithm complexity. Listed clustering and classification algorithms available within \pkg{reval} package from the \pkg{scikit-learn} and \pkg{hdbscan} libraries.}
\label{tab:cmplx} 
\end{table}

\begin{table}[t!]
\centering
\begin{tabular}{lccc}
\hline
Dataset & Samples & Features & Classes \\ \hline
handwritten digits & $1797$ & $64$ & $10$\\
yeast & $1484$ & $8$ & $10$\\
banknote & $1372$ & $4$ & $2$\\ 
biodegradation & $1055$ & $41$ & $2$\\
transfusion & $748$ & $4$ & $2$\\
breast cancer (WI) & $683$ & $9$ & $2$\\
urban land cover & $675$ & $147$ & $9$\\
climate & $540$ & $18$ & $2$\\
forest type & $523$ & $27$ & $4$\\
wholesale & $440$ & $7$ & $3$\\
movement & $360$ & $90$ & $15$\\
ionosphere & $351$ & $34$ & $2$\\
liver disorders & $345$ & $5$ & $2$\\
leaf & $340$ & $14$ & $30$\\
ecoli & $336$ & $7$ & $8$\\
glass & $214$ & $9$ & $6$\\
seeds & $210$ & $7$ & $3$\\
parkinsons & $195$ & $22$ & $2$\\
iris & $150$ & $4$ & $3$\\
\hline
\end{tabular}
\caption{Benchmark datasets from the UCI machine learning repository.}
\label{tab:dataset} 
\end{table}

\begin{table}
\centering
\resizebox{\columnwidth}{!}{%
\begin{tabular}{llllllllll}
\hline
&&&&&\multicolumn{4}{c}{Performance}\\
\cline{6-9}
\multicolumn{1}{c}{Dataset} & \multicolumn{1}{c}{Classes} &
\multicolumn{1}{c}{Clusters} &
\multicolumn{1}{c}{Best model} &
\multicolumn{1}{c}{Preprocessing} & \multicolumn{1}{c}{Validation stability (error)} & \multicolumn{1}{c}{test ACC} &\multicolumn{1}{c}{AMI} & \multicolumn{1}{c}{ACC} & \multicolumn{1}{c}{MCC}\\\hline

handwritten digits & $10$ & $10$ & KNN/KMeans & umap & $0.0$ & $0.99$ & $0.76$ & $0.82$ & $0.80$\\
yeast & $10$ & $7$ & \textbf{RF/Kmeans} & scaled+umap &  $0.05\ (0.01)$ & $0.94$ & $0.25$ & $0.39$ & $0.30$\\

banknote & $2$ & $2$ & SVM/HC & scaled+umap &  $0.003\ (0.006)$ & $1.00$ & $0.93$ & $0.99$ & $0.98$\\

biodegradation & $2$ & $2$ & KNN/KMeans & raw & $0.03\ (0.006)$ & $0.99$ & $-0.001$ & $0.58$ & $0.01$\\

transfusion & $2$ & $2$ & KNN/KMeans & umap & $0.0$ & $1.00$ & $0.005$ & $0.62$ & $0.10$\\

breast cancer (WI) & $2$ & $2$ & SVM/KMeans & raw & $0.03\ (0.01)$ & $0.99$ & $0.76$ & $0.96$ & $0.92$\\

urban land cover & $9$ & $3$ & \textbf{KNN/KMeans} & scaled+umap &  $0.006\ (0.003)$ & $0.95$ & $0.45$ & $0.41$ & $0.35$\\

climate & $2$ & $2$ & KNN/KMeans & scaled+umap & $0.20\ (0.04)$ & $0.50$ & $-0.005$ & $0.53$ & $0.02$\\

forest type & $4$ & $4$ & KNN/HC & raw & $0.35\ (0.11)$ & $0.88$ & $0.55$ & $0.79$ & $0.72$\\

wholesale & $3$ & $3$ & SVM/KMeans & umap & $0.07\ (0.03)$ & $0.97$ & $0.002$ & $0.37$ & $0.09$\\

movement & $15$ & $6$ & \textbf{KNN/HC} & umap & $0.05\ (0.02)$ & $0.90$ & $0.42$ & $0.32$ & $0.29$\\

ionosphere & $2$ & $2$ & SVM/KMeans & raw & $0.01\ (0.009)$ & $0.99$ & $0.21$ & $0.77$ & $0.52$\\

liver disorders & $2$ & $2$ & KNN/KMeans & umap & $0.11\ (0.04)$ & $0.92$ & $0.03$ & $0.62$ & $0.23$\\

leaf & $30$ & $3$ & \textbf{RF/KMeans} & scaled & $0.19\ (0.03)$ & $0.91$ & $0.22$ & $0.12$ & $0.11$\\

ecoli & $8$ & $2$ & \textbf{KNN/KMeans} & umap & $0.0$ & $1.00$ & $0.48$ & $0.62$ & $0.46$\\

glass & $6$ & $3$ & \textbf{KNN/KMeans} & scaled & $0.39\ (0.06)$ & $0.73$ & $0.35$ & $0.47$ & $0.35$\\

seeds & $3$ & $3$ & SVM/KMeans & raw & $0.05\ (0.03)$ & $0.89$ & $0.6$ & $0.89$ & $0.84$\\

parkinsons & $2$ & $2$ & KNN/KMeans & scaled+umap &  $0.02\ (0.01)$ & $0.96$ & $0.09$ & $0.63$ & $0.35$\\

iris & $3$ & $3/2$ (tr/ts) & RF/HDBSCAN & umap &  $0.33\ (0.19)$ & $0.88$ & $0.73$ & $0.67$ & $0.61$\\

\hline
\multicolumn{10}{l}{\footnotesize HC = hierarchical clustering; KNN = k-nearest neighbors; RF = random forest; SVM = support vector machine; umap = uniform manifold approximation and projection;}\\ \multicolumn{10}{l}{\footnotesize AMI = adjusted mutual information score; ACC = accuracy; MCC = Matthews correlation coefficient}
\end{tabular}
}
\caption{Best combinations of classifiers and clustering methods for benchmark datasets from the UCI machine learning repository. \pkg{reval} with grid search is applied. Normalized stability on validation with (error), and metrics are reported for performance evaluation. In bold results that failed in identifying the correct number of clusters.}
\label{tab:result}
\end{table}

\begin{table}
\centering
\resizebox{\columnwidth}{!}{%
\begin{tabular}{llllllll}
\hline
&&\multicolumn{3}{c}{Silhouette} & \multicolumn{3}{c}{Davies-Bouldin}\\
\cline{3-8}
\multicolumn{1}{c}{Dataset} & \multicolumn{1}{c}{Classes} &  \multicolumn{1}{c}{Clusters tr/ts} &  \multicolumn{1}{c}{Score tr/ts} & \multicolumn{1}{c}{AMI} & \multicolumn{1}{c}{Clusters tr/ts} &  \multicolumn{1}{c}{Score tr/ts} & \multicolumn{1}{c}{AMI}\\\hline
handwritten digits & $10$ & $10/10$ & $0.73/0.75$ & $0.76$ & $10/10$ & $0.39/0.36$ & $0.76$\\ 
yeast & $10$ & $4/4$ & $0.51/0.46$ & $0.25$ & $4/4$ & $0.65/0.72$ & $0.25$\\
banknote & $2$ & $2/2$ & $0.55/0.55$ & $0.93$ & $2/2$ & $0.72/0.72$ & $0.93$\\
biodegradation & $2$ & $2/2$ & $0.39/0.38$ & $-0.002$ & $2/2$ & $1.07/1.12$ & $-0.002$\\
transfusion & $2$ & $2/2$ & $0.45/0.46$ & $0.005$ & $2/2$ & $0.98/0.94$ & $0.005$\\
breast cancer (WI) & $2$ & $2/2$ & $0.59/0.61$ & $0.76$ & $2/2$ & $0.77/0.73$ & $0.76$\\
urban land cover & $9$ & $3/3$ & $0.60/0.55$ & $0.45$ & $3/3$ & $0.60/0.61$ & $0.45$\\
climate & $2$ & $2/2$ & $0.38/0.49$ & $-0.005$ & $2/2$ & $1.06/0.93$ & $-0.005$\\
forest type & $4$ & $4/4$ & $0.30/0.30$ & $0.55$ & $4/4$ & $1.19/1.15$ & $0.55$\\
wholesale & $3$ & $3/3$ & $0.54/0.54$ & $0.002$ & $3/3$ & $0.62/0.66$ & $0.002$\\
movement & $15$ & $6/6$ & $0.63/0.48$ & $0.42$ & $6/6$ & $0.52/0.73$ & $0.42$\\
ionosphere & $2$ & $2/2$ & $0.29/0.30$ & $0.21$ & $2/2$ & $1.51/1.50$ & $0.21$\\
liver disorders & $2$ & $2/2$ & $0.45/0.49$ & $0.03$ & $2/2$ & $0.86/0.78$ & $0.03$\\
leaf & $30$ & $3/3$ & $0.34/0.33$ & $0.22$ & $3/3$ & $1.09/1.09$ & $0.22$\\
ecoli & $8$ & $2/2$ & $0.75/0.72$ & $0.48$ & $2/2$ & $0.33/0.37$ & $0.48$\\
glass & $6$ & $3/3$ & $0.41/0.40$ & $0.35$ & $3/3$ & $1.34/0.96$ & $0.35$\\
seeds & $3$ & $3/3$ & $0.49/0.44$ & $0.66$ & $3/3$ & $0.69/0.82$ & $0.66$\\
parkinsons & $2$ & $2/2$ & $0.61/0.53$ & $0.09$ & $2/2$ & $0.55/0.73$ & $0.09$\\
iris & $3$ & $4/2$ & $0.76/0.93$ & $0.73$ & $4/2$ & $0.37/0.09$ & $0.73$\\
\hline
\multicolumn{8}{l}{\footnotesize AMI = adjusted mutual information score}\\
\end{tabular}
}
\caption{Internal measure performance for the benchmark datasets from the UCI machine learning repository. Clustering solutions are based on best silhouette and Davies-Bouldin scores. Best solution is defined, on both training and test sets, as the one maximizing and minimizing the internal measures, respectively. Adjusted mutual information on test set is reported for performance evaluation.}
\label{tab:resultint} 
\end{table}

\clearpage


\section*{Supplementary material}

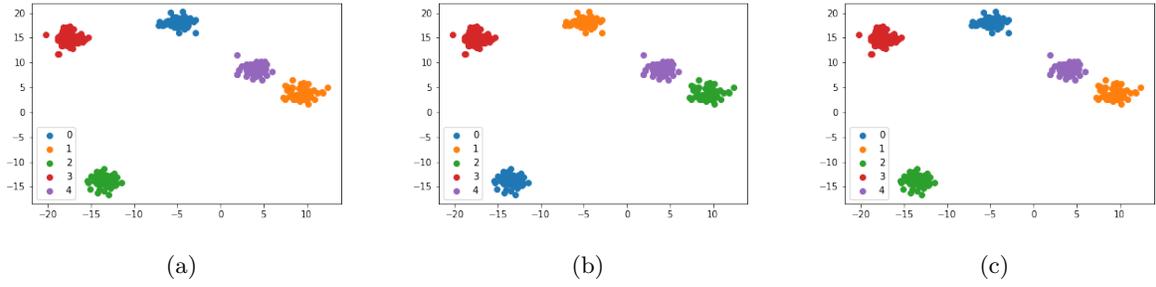

Supplementary Figure 3: Blobs dataset test set visualization. True blob labels (a), best clustering solution labels (b), and clustering labels obtained after permutation (c).

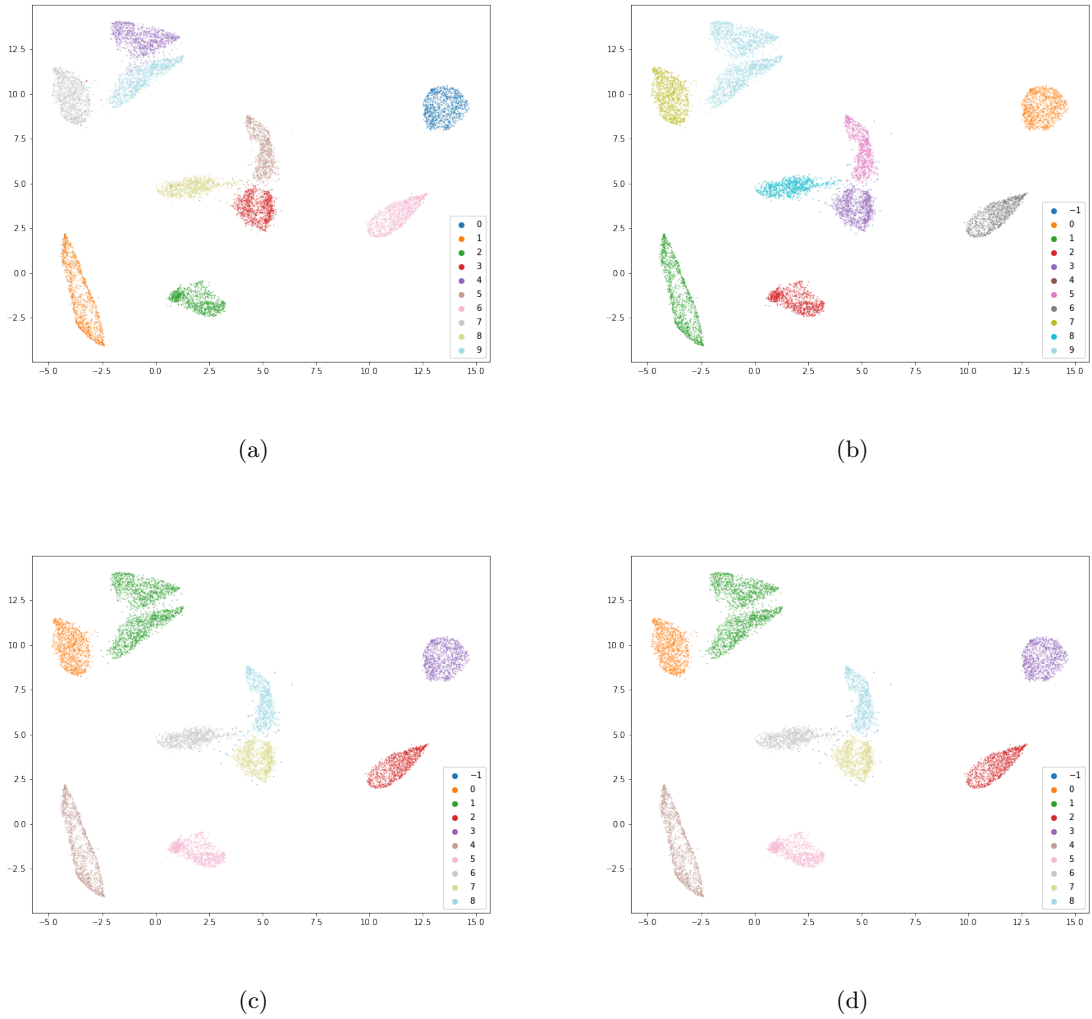

Supplementary Figure 4: MNIST test set UMAP visualization. True (a); relative validation (b); silhouette score (c); Davies-Bouldin index (d) labels.

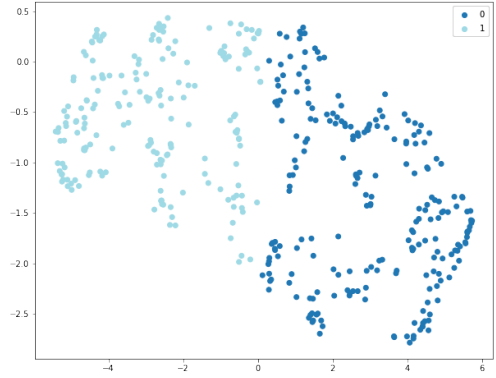

(a)

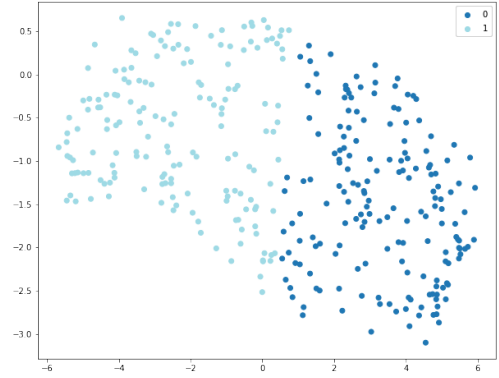

(b)

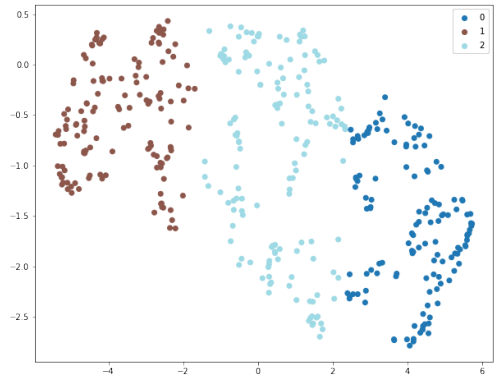

(c)

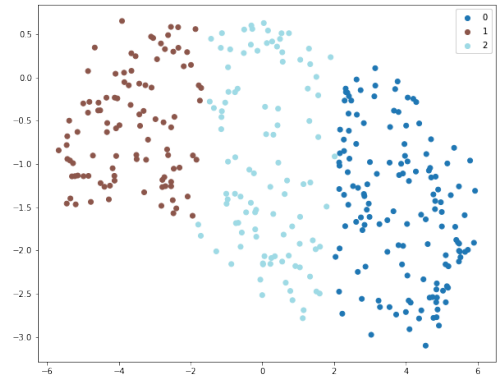

(d)

Supplementary Figure 5: Clustering solution comparisons for the national database for autism research (NDAR) dataset. Data is preprocessed with uniform manifold approximation and projection (UMAP) for visualization. 2-cluster solution for NDAR training (a) and test (b) sets; 3-cluster solution for NDAR training (c) and test (d) sets.

## Blobs dataset

Generate the dataset with 1000 samples and 2 features and visualize the blobs (see Supplementary Figure 9).

```
data = make_blobs(1000, 2, centers=5,

                  center_box=(-20, 20),

                  random_state=42)


plt.figure(figsize=(6, 4))

for i in range(5):

    plt.scatter(data[0][data[1]==i][:, 0],

                data[0][data[1]==i][:, 1],

                label=i, cmap='tab20')

plt.title("Blobs dataset")

plt.show()
```

Create training and test sets (30% of data) and initialize classifier/classification algorithms.

```
X_tr, X_ts, y_tr, y_ts = train_test_split(data[0],

                                          data[1],

                                          test_size=0.30,

                                          random_state=42,

                                          stratify=data[1])

classifier = KNeighborsClassifier(n_neighbors=15)

clustering = KMeans()
```

Run stability-based relative validation, plot repeated cross validation performance metrics and evaluate the best solution on the test set.

```
findbestclust = FindBestClustCV(nfold=2,

                                nclust_range=list(range(2, 7, 1)),

                                s=classifier,

                                c=clustering,

                                nrand=10,

                                n_jobs=1)

metrics, nbest = findbestclust.best_nclust(X_tr, iter_cv=10, strat_vect=y_tr)

plot_metrics(metrics, prob_lines=False)
```

```
out = findbestclust.evaluate(X_tr, X_ts, nclust=nbest)
print(f"Validation stability: {metrics['val'][nbest]}")
```

To compare predicted and true labels first we have to find the best permutation for the clustering labels, i.e., the one that minimizes the misclassification error. This because both classifier and clustering labels, and true and clustering labels are not necessarily in natural correspondence. Supplementary Figure 10 compares true labels (a) with clustering labels (b). It is straightforward to note that the label ordering does not match. After running the Kuhn-Munkres algorithm the correspondence between true and clustering labels is preserved (see Supplementary Figure 10c).

```
perm_lab = kuhn_munkres_algorithm(y_ts, out.test_cllab)


print(f"Best number of clusters: {nbest}")
print(f'AMI (true labels vs predicted labels) for test set = '
      f'{adjusted_mutual_info_score(y_ts, out.test_cllab)}')
print('\n\n')
print("Metrics from true label comparisons on test set:")
class_scores = compute_metrics(y_ts, perm_lab, perm=False)
for k, val in class_scores.items():
    if k in ['ACC', 'MCC']:
        logging.info(f"{k}, {val}")
logging.info("\n\n")
```

The following code selects the best number of clusters based on silhouette and Davies-Bouldin scores for k-means clustering applied to both training and test sets and returns the AMI scores.

```
print("Silhouette score based selection")
sil_score_tr, sil_best_tr, sil_label_tr = select_best(X_tr, clustering,
                                                      silhouette_score,
                                                      select='max',
                                                      nclust_range=list(range(2, 7, 1)))
sil_score_ts, sil_best_ts, sil_label_ts = select_best(X_ts, clustering,
                                                      silhouette_score,
                                                      select='max',
```

```
                                              nclust_range=list(range(2, 7, 1)))

sil_eval = evaluate_best(X_ts, clustering, silhouette_score, sil_best_tr)

print(f"Best number of clusters (and scores) for tr/ts independent runs:
      {sil_best_tr}({sil_score_tr})/{sil_best_ts}({sil_score_ts})")
print(f"Test set evaluation {sil_eval}")
print(f'AMI (true labels vs clustering labels) training = '
            f'{adjusted_mutual_info_score(y_tr,
              kuhn_munkres_algorithm(y_tr, sil_label_tr))}')
print(f'AMI (true labels vs clustering labels) test = '
            f'{adjusted_mutual_info_score(y_ts,
              kuhn_munkres_algorithm(y_ts, sil_label_ts))}')
print('\n\n')


print("Davies-Bouldin score based selection")
db_score_tr, db_best_tr, db_label_tr = select_best(X_tr, clustering,
                                              davies_bouldin_score,
                                              select='min',
                                              nclust_range=list(range(2, 7, 1)))
db_score_ts, db_best_ts, db_label_ts = select_best(X_ts, clustering,
                                              davies_bouldin_score,
                                              select='min',
                                              nclust_range=list(range(2, 7, 1)))


db_eval = evaluate_best(X_ts, clustering, davies_bouldin_score, db_best_tr)


print(f"Best number of clusters (and scores) for tr/ts independent runs:
      {db_best_tr}({db_score_tr})/{db_best_ts}({db_score_ts})")
print(f"Test set evaluation {db_eval}")
print(f'AMI (true labels vs clustering labels) training = '
            f'{adjusted_mutual_info_score(y_tr,
            kuhn_munkres_algorithm(y_tr, db_label_tr))}')
```

6

```
print(f'AMI (true labels vs clustering labels) test = '
                f'{adjusted_mutual_info_score(y_ts,
                kuhn_munkres_algorithm(y_ts, db_label_ts))}')
print('\n\n')
```

Plot true labels and clustering labels for test set.

```
plt.figure(figsize=(6, 4))
plt.scatter(X_ts[:, 0], X_ts[:, 1],
            c=y_ts, cmap='tab20')
plt.title("Test set true labels")
plt.show()


plt.figure(figsize=(6, 4))
plt.scatter(X_ts[:, 0], X_ts[:, 1],
            c=perm_lab, cmap='tab20')
plt.title("Test set clustering labels")
plt.show()
```


\end{document}